\documentclass{Interspeech2024}

\interspeechcameraready

\title{Improving endpoint detection in end-to-end streaming ASR \\ for conversational speech}

\name[affiliation={1,2}]{Anandh}{C}
\name[affiliation={1}]{Karthik Pandia}{Durai}
\name[affiliation={1}]{Jeena}{Prakash}
\name[affiliation={1}]{Manickavela}{Arumugam}
\name[affiliation={3}]{Kadri}{Hacioglu}
\name[affiliation={1}]{S. Pavankumar}{Dubagunta}
\name[affiliation={3}]{Andreas}{Stolcke}
\name[affiliation={1}]{Shankar}{Venkatesan}
\name[affiliation={1}]{Aravind}{Ganapathiraju}

\address{
  $^1$Uniphore Software Systems, India\\
  $^2$Indian Institute of Technology Madras, India
  $^3$Uniphore Software Systems, USA}
\email{ee22s076@smail.iitm.ac.in, karthik.pandia@gmail.com}

\keywords{endpointing, latency, E2E ASR, RNN-T, Zipformer, transducer}

\begin{document}

\maketitle

\begin{abstract}
ASR endpointing (EP) plays a major role in delivering a good user experience in products supporting human or artificial agents in human-human/machine conversations. Transducer-based ASR (T-ASR) is an end-to-end (E2E) ASR modelling technique preferred for streaming. A major limitation of T-ASR is delayed emission of ASR outputs, which could lead to errors or delays in EP. Inaccurate EP will cut the user off while speaking, returning incomplete transcript while delays in EP will increase the perceived latency, degrading the user experience. We propose methods to improve EP by addressing delayed emission along with EP mistakes. To address the delayed emission problem, we introduce an end-of-word token at the end of each word, along with a delay penalty. The EP delay is addressed by obtaining a reliable frame-level speech activity detection using an auxiliary network. We apply the proposed methods on Switchboard conversational speech corpus and evaluate it against a delay penalty method.

\end{abstract}

\section{Introduction}
\label{sec:intro}
Speech endpointing, i.e., the detection of end-of-speech in a turn, is an important aspect of conversational speech recognition. Two-party human-to-human/machine conversations in an industrial setting, such as a customer speaking to a call center agent, or a customer speaking to a voice bot, require precise endpointing such that the live transcription is organized into correct turns during the conversation, which in-turn leads to timely and accurate extraction of further information. In the case of bot-driven calls, effective endpointing improves task completion rate. Streaming ASR systems should not only achieve a low word error rate (WER) but also exhibit low latency for a good user experience. To achieve this, the choice of the model architecture and the choice of endpointing approach plays an important role.

End-to-end (E2E) ASR models have become the de facto standard ASR models for industrial use cases. Connectionist temporal classification (CTC) \cite{graves2014towards}, recurrent neural network transducer (RNN-T) \cite{graves2012sequence}, and attention-based encoder-decoder (AED) \cite{chorowski2014end} are the main E2E ASR techniques used. An encoder is used in all three, while the approaches differ in their decoding methods.
CTC assumes conditional independence of the output symbols in the loss function, which is seen as a limitation of the technique.
AED uses an encoder in output, and monotonic cross-attention \cite{raffel2017online} is used to obtain the alignment between inputs and outputs.
Generally, for AED-based ASR, a CTC model is utilised as an auxiliary model to obtain a good alignment between input frames and output symbols \cite{kim2017joint}. 
In RNN-T, the current output is conditioned on the previous outputs. An autoregressive network is used on the output tokens. The representation obtained from this network, together with the encoder representation, is used in a joiner network that predicts the next output token. More recently, RNN encoders have been replaced by transformer-based encoders such as Conformer \cite{gulati2020conformer} and Zipformer\cite{yao2023zipformer} . We refer to such an architecture as transducer-based ASR (T-ASR).
In streaming use cases, the technique should not only support streaming capabilities, but also exhibit low latency for a good user experience. Monotonic attention has been used to support streaming in AED. However, T-ASR has been shown to outperform AED in terms of latency and throughput \cite{kim2021comparison}. T-ASR is a widely used streaming E2E ASR technique. T-ASR still suffers from latency problems due to the delayed emission of output tokens~\cite{li2020towards,yu2021fastemit,kang2023delay}.
\cite{mahadeokar2021alignment} constrained the training alignments to be closer to a predefined reference obtained through forced alignment, thereby limiting the token emission delay. Similarly, \cite{kim21j_interspeech} used a training loss that encouraged advanced predictions of tokens as compared to a forced-aligned reference. \cite{shinohara22_interspeech} penalised late prediction of tokens by defining an expected latency for each frame. Work such as \cite{yu2021fastemit,kang2023delay} use losses that encourage non-blank token predictions.

User latency is affected by delayed emission and delayed endpointing, though delayed endpointing can also be an effect of delayed emission.
In the T-ASR framework, to address latency and endpointing jointly, \cite{li2020towards,lu2022endpoint} used a special token to detect the end of a sentence (EOS) and by penalising its late prediction in the training loss. In \cite{fan2023towards}, a binarised (speech/non-speech) state sequence from the joiner output was used to achieve a forced alignment at the frame level. The corresponding probabilities were obtained using an independent long-short-term memory (LSTM) network, which used encoder outputs as its inputs. Here, the last senone state is considered as an EOS marker. 
\cite{bijwadia2023unified} also used two parallel networks, one for ASR prediction and another for endpointing. The endpoint model was trained to predict endpoints using either acoustic features or latent representations from a shared layer. In addition to using the \textit{final silence} state to detect end of a query (EOQ), the authors proposed using a special EOQ token in the transcripts, thereby detecting EOQ both using acoustics and text.

In this work, we make three contributions: (A) Using encoder embeddings and a light-weight network to reliably estimate the endpoints; (B) using an end-of-word token to avoid word fragments while doing aggressive endpointing; (C) novel endpointing strategies that improve the trade-off between latency and word error rate.
Existing endpointing methods for E2E ASR are typically evaluated on voice search tasks or on the Librispeech \cite{panayotov2015librispeech} read speech dataset. We evaluate our methods on spontaneous speech  taken from the  Switchboard conversational telephone speech corpus~\cite{godfrey1992switchboard}. A Zipformer transducer T-RNN \cite{yao2023zipformer} is used for all our experiments.

The rest of the paper is organized as follows. Section~\ref{sec:prop} describes the proposed methodology for endpoint detection in detail. The experimental setup, results, and analysis are found in Section~\ref{sec:exp}. Section~\ref{sec:conc} offers conclusions.

\section{Proposed methods}
\label{sec:prop}

We now describe the problem of endpointing delay and the proposed methods to address it.

\subsection{Endpointing delay in transducer-based ASR}


In ASR systems, the trailing silence (non-speech following a speech region) is generally used to detect endpoint, though contextual information from the preceding speech region and conversation can also help predict ends of speaker turns \cite{ferrer2002speaker,WangEtAl:icassp2024}.
In T-ASR, non-speech is subsumed by the blank token. The training loss function mandates that a single non-blank token be emitted for every occurrence of a sub-word unit in speech. This means that the majority of the frame-level predictions are blank tokens around these non-blank emissions, in both speech and non-speech regions. Thus, to detect non-speech, the model provides no alternative but to search for the timestamps---or regions---that have no ``occasional'' non-blank tokens emitted. This is achieved by setting a threshold on the number of consecutive blank tokens emitted, above which an endpoint is declared~\cite{shangguan21_interspeech}. This formulates endpoint prediction as a detection problem, where a too-low threshold can lead to a false positive (premature endpoint), and a too-high threshold can lead to missed, as well as delayed, predictions of the endpoints.


The three issues can be addressed by using a reliable estimate for trailing silence and prompt emission of non-blanks. In this work, we propose to use a separate neural network speech detection (voice activity, or VAD) model to classify a frame as speech or nonspeech. The output of the model is used to compute trailing nonspeech duration, which can then be used to trigger an endpoint.
Crucially, the VAD model is tightly coupled with the T-ASR model, using the output of the audio encoder and feeding into the ASR output decision layer.
We show that using the transcription network encoding to detect an endpoint leads to a reliable classification of frames as speech/nonspeech. 
Additionally, an end-of-word (EOW) token along with a delay penalty \cite{kang2023delay} during training leads to prompt emission of non-blank at the word endings. Using EOW facilitates endpointing with low latency.

\begin{figure}[tb]
  \centering
  \includegraphics[width=\linewidth]{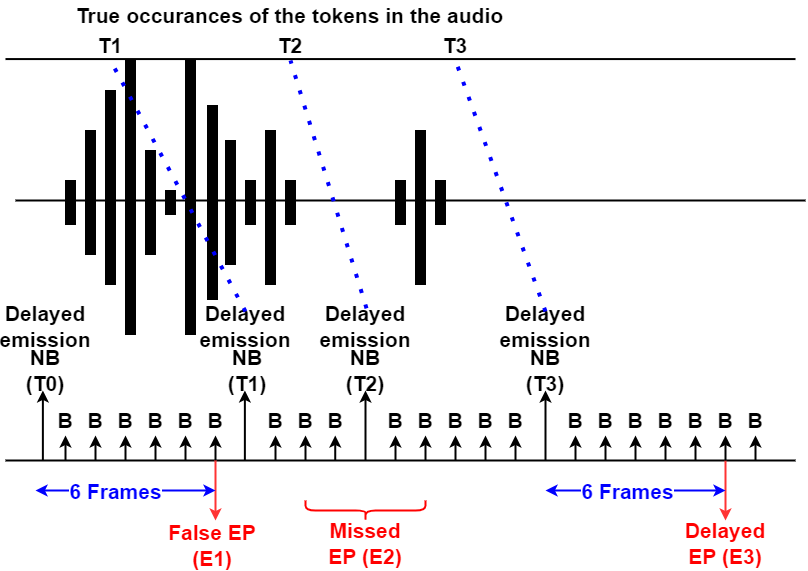}
  \caption{Illustration of false EP, missed EP, and delayed EP while using blank symbols for endpointing. Here, 6 frames of blank tokens are required to trigger an endpoint.}
  \label{fig:endpoint_illustration}
\end{figure}

As illustrated in Figure~\ref{fig:endpoint_illustration}, using the blank token can lead to false, missed, or delayed endpoint detection. NB and B in Figure~\ref{fig:endpoint_illustration} represents the emission of non-blank or blank, respectively, for each frame. A total of 6 frames with blank emissions is required to trigger an endpoint. $T_1$, $T_2$, and $T_3$ are the timestamps where the acoustics corresponding to non-blank emissions are present. Endpoint $E_1$ is a false endpoint due to the delayed emission of NB from $T_1$. Similarly, delayed emission of non-blank from $T_2$ leads to missing of expected endpoint $E_2$. Delayed emission of the non-blank from $T_3$ leads to delayed endpointing ($E_3$). We propose methods to address these problems.

\subsection{VAD network for reliable trailing silence estimation}

\begin{figure}[tb]
  \centering
  \includegraphics[width=0.8\linewidth]{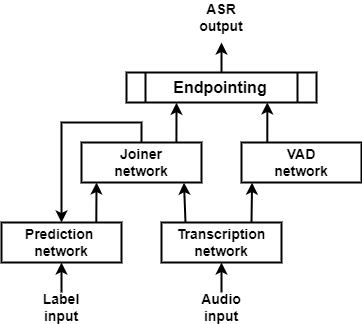}
  \caption{Block diagram of proposed endpointing method}
  \label{fig:block_proposed}
\end{figure}

The block diagram of the proposed method is shown in Figure~\ref{fig:block_proposed}. A VAD network is introduced to classify frames as speech or nonspeech (silence). The decision from the VAD network is used to trigger endpoints.
Output from the transcription encoder network (Zipformer encoder) is used as input representation for VAD. A feed-forward neural network is used as a classifier and the output is binary. An endpoint is detected right after a contiguous silence of $x$\,ms. The decision from the VAD classifier is used to obtain the duration of silence. 

\begin{figure}[tb]
  \centering
  \includegraphics[width=\linewidth]{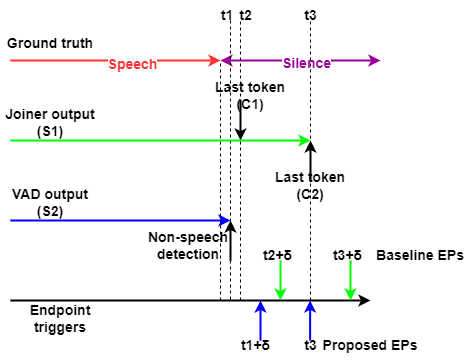}
  \caption{C1 and C2 are the possible endpointing cases where the joiner output is emitted before ($t_2$) or after ($t_3$) the VAD decision ($t_1+\delta$), respectively. Here $\delta$ is the trailing silence (after speech ended) required to trigger an endpoint. $t_1+\delta$ and $t_3$ are the endpoints of the proposed method; $t_1+\delta$ and $t_3+\delta$ are the endpoints of the baseline for the cases C1 and C2, respectively.}
  \label{fig:problem_illustration}
\vspace{-2mm}
\end{figure}

A limitation of this approach becomes apparent when the joiner non-blank symbol output is delayed, but is expected to appear after the endpoint trigger. Endpointing before seeing the last token of a word can lead to word fragments or deletions. A synchronisation mechanism between the VAD output and joiner output is required to handle this case. Figure~\ref{fig:problem_illustration} illustrates the problem. 

\subsection{EOW token to mitigate word breaks}

In Figure~\ref{fig:problem_illustration}, the event at $t_1$ corresponds to the first non-speech detection from the VAD network. An endpoint is detected after $\delta$ at $t_1+\delta$. $t_2$ corresponds to a timestamp of the joiner output of the last token. In case (C1), if the joiner output for the last token comes before $t2$ and the endpoint event $t1+\delta$, then the endpoint decision can be taken as it is. In case 2 (C2), if the joiner output for the last token comes after $t_2$ and the endpoint event $t_1+\delta$, then deciding an endpoint at $t_1+\delta$ can lead to an error in the last word. To address this issue, an end-of-word symbol (EOW) is introduced in ASR model training. Each word in the training data is followed by a special EOW symbol. An endpoint is finalised only when the last output from the joiner is EOW. Otherwise, the endpoint is made after the occurrence of the EOW. In this method, the problem of delayed emission of non-blank persists. Delay of the EOW can lead to delayed or missed endpoint detection. To address this problem, the model is trained with a delay penalty as proposed in ~\cite{kang2023delay}.
Now, the possible endpointing strategies are
\begin{itemize}
    \item Trailing silence (TS) duration condition (TS rule)
    \item When EOW is seen during TS aggregation (EOW rule)
    \item Both EOW and TS conditions are met
\end{itemize}
These endpointing methods are analysed in Section~\ref{sec:exp}.


\section{Datasets, experiments, and results}
\label{sec:exp}

\subsection{Dataset description and modelling framework}
\label{ssec:dataset}

We use the Switchboard corpus \cite{godfrey1992switchboard} in our experiments since it contains realistic spontaneous, human-human speech, as could be expected from natural human-machine dialog, and unlike pre-planned or read speech.
The test set from \cite{stolcke2000dialogue} was used. The remaining data from the corpus is used to split train and validation sets. Table~\ref{tab:dataset} shows the details of the train, validation, and test partitions. These datasets are used for comparing VAD models, estimating the operating point, and ASR training. To measure endpointing and its effect on WER, decoding has to be at the call level. Therefore, the segment-level transcripts for a call are combined to obtain call-level transcripts. 

\begin{table}[h]
  \caption{Switchboard corpus partitions used in experiments}
  \label{tab:dataset}
  \centering
  \begin{tabular}{l|c|c|c}
    \toprule
    \ & \textbf{\# calls} & \textbf{Duration (h)} & \textbf{\# speakers}   \\
    \midrule
    \textbf{Train} & 2353 &  305.2 & 359 \\
    \textbf{Validation} &  132 & 8.9 & 33 \\
    \textbf{Test} & 19 & 3.1 & 8 \\
    \bottomrule
  \end{tabular}
\end{table}

Icefall \footnote{https://github.com/k2-fsa/icefall} is used for T-ASR experiments. Zipformer is used as the encoder. The acoustic inputs are 80-dimensional Mel filter bank features with a 25\,ms frame size and 1\,ms frame shift. The encoder output is obtained for every 40\,ms. 
Endpoint detection is an event detection task. The reference contains true events, while putative events are hypothesised by the system. Events from the reference and hypotheses are aligned to calculate hits and misses. Finally, precision and recall metrics are calculated using the \textit{sed\_eval} toolbox \cite{mesaros2016metrics}. A tolerance of 200\,ms from the reference location is applied to decide whether an event is hit or miss. A range of trailing silence duration of {200, 400, 600, and 800}\,ms are considered for endpoint triggering. However, the main results are determined using the 200\,ms configuration. 

\subsection{encNET and melNET for VAD}
\label{ssec:vadnet}
The encoder output is used to train a feed-forward neural network (encNET) for voice activity detection. 
The reference annotations provided with the Switchboard corpus does not reflect cross-talk that was pervasive in the data. The VAD we propose is not designed to eliminate cross-talk. Therefore we use an external VAD that gives reference information that is more realistic for a single-channel VAD task.
Specifically, we use the Kaldi voice activity detection model trained on Fisher English corpus (ASpIRE model) \cite{ryant2021third}. We found that the detection error rate (DER) of the ASpIRE VAD model is high (\textbf{$0.426$}) with respect to the reference timestamps. This justifies the usage of the ASpIRE VAD as a teacher model to train the encNET. The usage of the VAD model is also pertinent in scenarios where reference timestamps are not available.


\begin{figure}[h]
  \centering
   \includegraphics[width=\linewidth]{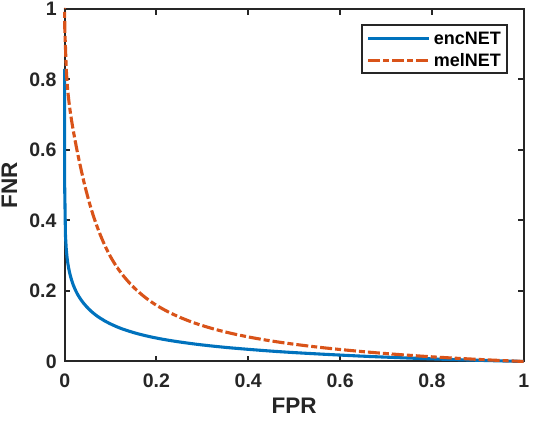}
  \caption{Detection error trade-off curves for encNET and melNET. The EER for encNET and melNET are 0.105 and 0.182, respectively.}
  \label{fig:det}
\end{figure}    

The encNET is a feed-forward neural network with two hidden layers, trained to classify speech or nonspeech at the frame level. 
The encoding from the Zipformer encoder is used as input to this model. This means the encNET needs to be retrained for a new ASR model (different language/dataset) since the input representations to encNET are specific to the encoder. Therefore, we also evaluated VAD based on Mel filter bank features instead of encoder embedding. The VAD network thus trained is denoted as melNET. We compare encNET and melNET to understand the advantage of using ASR-induced encoder embeddings in frame-level VAD decisions. A 100\,h subset of the Switchboard training data is used to train both the networks. The validation set is used to compare the models on the frame-wise speech/non-speech classification task. Figure~\ref{fig:det} shows the detection error trade-off (DET) curve. From the plot, we clearly see that encNET is superior to melNET. 

Changing the acceptance threshold for speech vs.\ nonspeech yields different operating points for VAD. Since both false positive rate and false negative rate are equally important to reliably detect endpoints, we use the operating point corresponding to the equal error rate (EER) for evaluation. EER for encNET and melNET are $0.105$ and $0.182$, respectively.

\subsection{Endpointing using encNET and melNET}
\label{ssec:endpointing}

The results reported in this section are based on the trailing silence (TS) rule for endpointing. Table~\ref{tab:p_r_bp_comparison} shows the precision, recall, F1 score, and WER for four systems. The baseline (vanilla) approach is endpointing using blank symbol aggregation for estimating trailing silence duration. The improvement that delay penalty (DP) as proposed in \cite{kang2023delay} brings over the vanilla model is shown in the second row. A delay penalty of $0.0004$ worked best in our experiment based on the WER criterion. Endpointing with the proposed melNET and encNET VAD systems using the TS rule are shown in the last two rows. The oracle WER of the model without any endpointing is $21.92$\%.

\begin{table}[tb]
  \caption{Precision, recall, F1 score, and WER for the proposed VAD methods (melNET and encNET) and blank-based EP methods (vanilla and delay penalty (DP)\cite{kang2023delay})}
  \label{tab:p_r_bp_comparison}
  \centering
  \begin{tabular}{l|c|c|c|c}
    \toprule
    \  & \textbf{Prec.} & \textbf{Recall} & \textbf{F1}  & \textbf{WER}   \\
    \midrule
    \textbf{Oracle (without EP)} & - & - & - & 21.92 \\ 
    \textbf{Blank-based vanilla} & 33.6 & 76.2 & 46.6 & 24.4 \\
    \textbf{Blank penalty DP} & 34.1 & 77.4 & 47.3 & 24.5 \\  \hline
    \rule{0pt}{2ex}\textbf{melNET} & 	54.0 & 83.7 & 65.6 & 22.9 \\
    \textbf{encNET} & 	74.9 & 82.8 & 78.6 & 23.0 \\
    \bottomrule
  \end{tabular}
\end{table}

From Table~\ref{tab:p_r_bp_comparison}, we see that having an external VAD helps reduce the endpointing mistakes significantly, leading to improved precision and recall. Specifically, high precision of melNET (33.6\% to 54.0\%) leads to a decrease in false triggers, thereby improving the WER (from 22.4\% to 22.9\%). The much higher precision of encNET leads to missing of endpoints, though not affecting the WER much (22.9\% to 23.0\%), or about 1\% absolute higher than the WER obtained with out endpointing ($21.92$\%).

\subsection{EOW for WER improvement}

Though the prediction of frame-level silence is good, the output from T-ASR is still delayed. We use the delay penalty method proposed in \cite{kang2023delay} to reduce the emission delay. In addition to adding a delay penalty during training, the EOW token is introduced to explicitly indicate the end of a word (EOW rule). In EOW rule, a minimum trailing silence followed by the EOW tag in the ASR hypothesis triggers an endpoint. Adding the delay penalty during model training along with the EOW rule should encourage fast emission of EOW. To achieve higher precision, both EOW and TS rules are applied to trigger an EP. Table~\ref{tab:p_r_bp_eow} shows the results using TS rule, EOW rule, and the combined rules.

\begin{table}[h]
  \caption{Results for models trained with blank penalty \cite{kang2023delay} and end-of-word (EOW) token for different endpointing strategies. encNET is used for trailing silence (TS) aggregation}
  \label{tab:p_r_bp_eow}
  \centering
  \begin{tabular}{l|c|c|c|c}
    \toprule
    \textbf{EP strategies}  & \textbf{Precision} & \textbf{Recall} & \textbf{F1}  & \textbf{WER}   \\
    \midrule
    Blank-based & 47.7	& 71.5 & 57.3 & 25.6 \\ \hline
    TS & 75.4	& 83.0 & 79.0 &	23.7 \\
    EOW & 81.4 & 65.6 & 72.6	& 21.4 \\
    TS and EOW & 87.5 & 62.7 & 73.0 & 21.7 \\
    \bottomrule
  \end{tabular}
\end{table}

EOW-based endpointing improves the precision significantly while a degradation in recall is observed. This increase in precision improves the WER significantly, achieving oracle WER for this model (21.4\%). EP using both TS and EOW condition leads to a slight degradation in WER while marginally improving the F1 score. 

%


\begin{figure}[h]
  \centering
  \includegraphics[width=\linewidth]{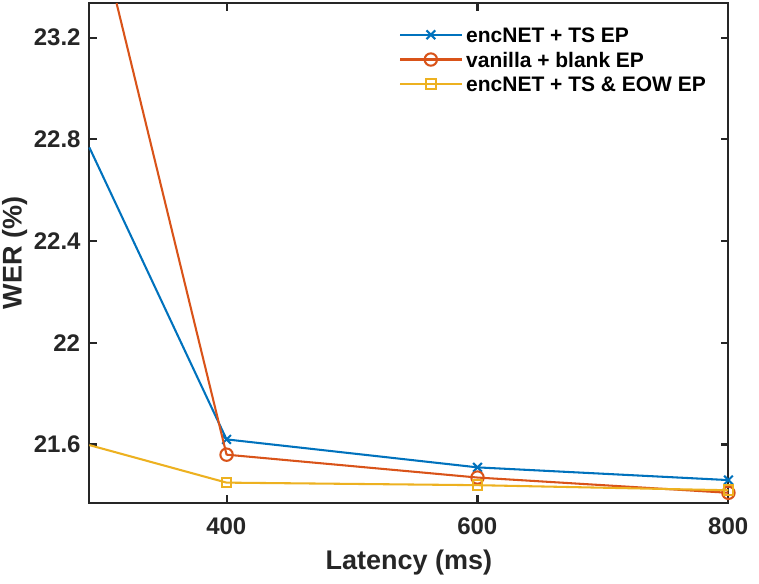}
\vspace{-0.5cm}
  \caption{Latency vs WER for different systems for various trailing silence configurations (400ms, 600ms, and 800ms)}
  \label{fig:latency_wer}
\end{figure}  

In the above results, endpointing was triggered for a trailing silence of 200\,ms. A more thorough comparison is possible by plotting the WER of proposed methods for various trailing silence values: 400, 600, and 800\,ms as shown in Figure~\ref{fig:latency_wer}, showing the trade-off between latency and WER. At around 600\,ms latency, the WER of the various systems saturates with WER close to that of oracle. However, an aggressive endpointing to achieve a latency of 200\,ms leads to significant degradation in the WER when the endpointing is only based on trailing silence.

\section{Conclusion}
\label{sec:conc}

We have proposed methods to improve endpointing in end-to-end transducer-based ASR systems. A separate speech detector network operates in parallel with the ASR decoder to determine speech/non-speech at the frame-level. We show that ASR encoder embedding can be reused for VAD for better results. Second, we introduce an end-of-word token in ASR training, that, when combined with the proposed endpointing logic avoids word fragmentation while keeping the latency low. Using the proposed methods, we were able to achieve high F1 score in detecting endpoints without any degradation from oracle WER.
\newpage

\bibliographystyle{IEEEtran}
\bibliography{mybib}

\end{document}